\documentclass[10pt,twocolumn]{article} 
\usepackage{simpleConference}
\usepackage{times}
\usepackage{graphicx}
\usepackage{amssymb}
\usepackage{url,hyperref}

\usepackage{amsmath,amsthm}
\usepackage{amsfonts}
\usepackage{dsfont}
\usepackage{cases}
\usepackage{bm}
\usepackage{algorithm}
\usepackage{algorithmic}
\usepackage{graphicx}
\usepackage{diagbox}
\usepackage{multirow}
\usepackage{xcolor}
\usepackage{float}
\usepackage[caption=false]{subfig}
\usepackage[numbers,sort&compress]{natbib}
\usepackage{longtable,lscape}
\usepackage{array}
\usepackage{booktabs}
\usepackage{threeparttable}
\usepackage{dcolumn}

\begin{document}

\title{Hierarchy Neighborhood Discriminative Hashing for An Unified View of Single-Label and Multi-Label Image retrieval}

\author{Lei Ma, Hongliang Li, Qingbo Wu, Fanman Meng and King Ngi Ngan \\
\\
School of Information and Communication Engineering \\
University of Electronic Science and Technology of China \\
\today
\\
\\
}

\maketitle
\thispagestyle{empty}

\begin{abstract}
Recently, deep supervised hashing methods have become popular for large-scale image retrieval task. To preserve the semantic similarity notion between examples, they typically utilize the pairwise supervision or the triplet supervised information for hash learning. However, these methods usually ignore the semantic class information which can help the improvement of the semantic discriminative ability of hash codes. In this paper, we propose a novel hierarchy neighborhood discriminative hashing method. Specifically, we construct a bipartite graph to build coarse semantic neighbourhood relationship between the sub-class feature centers and the embeddings features. Moreover, we utilize the pairwise supervised information to construct the fined semantic neighbourhood relationship between embeddings features. Finally, we propose a hierarchy neighborhood discriminative hashing loss to unify the single-label and multi-label image retrieval problem with a one-stream deep neural network architecture. Experimental results on two large-scale datasets demonstrate that the proposed method can outperform the state-of-the-art hashing methods.
\end{abstract}

\section{Introduction}
Hashing \cite{tmmMaLEBC,nipsWeissSH,npMaGLSPH,jvcirMaMREH,vcipMaMDLH,pamiGongITQ,ijcaiLiDPSH,nipsLiDSDH} has been paid attention by lots of researchers for large-scale image retrieval in recent years. The goal of hashing is to transform the multimedia data from the original high-dimensional space into a compact binary space while preserving data similarities. The const or sub-linear search speed can be achieved via Hamming distance measurement, which is performed by using XOR and POPCNT operations on modern CPUs or GPUs. The efficient storage and search make hashing technology popular for large-scale multimedia retrieval.

Generally, we can divide existing hashing approaches into two categories: data-independent and data-dependent hashing methods. Data-independent hashing methods \cite{vldbGionisLSH} map the data points from the original feature space into a binary-code space by using random projections as hash functions. These methods provide theoretical guarantees for mapping the nearby data points into the same hash codes with high probabilities. However, they need long binary codes to achieve high precision. Data-dependent hashing methods (i.e., learning to hash methods) \cite{tmmMaLEBC,nipsWeissSH,npMaGLSPH,pamiGongITQ,jvcirMaMREH,cvprLiuKSH,cvprShenSDH,ijcaiLiDPSH,ijcaiYaoDSRH,accvWangDTSH,nipsLiDSDH} learn hash functions and compact binary codes from training data. They can be further divided into unsupervised hashing methods \cite{nipsWeissSH,pamiGongITQ,aaaiKangCOSDISH,jvcirMaMREH,tmmMaLEBC} and supervised hashing methods \cite{cvprLiuKSH,pamiWangSSH,cvprShenSDH}, based on whether or not the semantic (label) information is used. In many real applications, supervised hashing methods demonstrate superior performance over unsupervised hashing methods. Recently, deep learning based hashing methods \cite{tipJiangDDSH,ijcaiLiDPSH,ijcaiYaoDSRH,accvWangDTSH,nipsLiDSDH,npMaGLSPH} demonstrate superior performance over these traditional hashing methods \cite{nipsWeissSH,pamiGongITQ,cvprLiuKSH,jvcirMaMREH,tmmMaLEBC}. The main reason is that deep hashing methods can perform simultaneous feature learning and hash-code learning in an end-to-end framework. Existing deep supervised hashing methods \cite{aaaiCaoDHN, cvprLaiDNNH, aaaiXiaCNNH, ijcaiLiDPSH,ijcaiYaoDSRH,accvWangDTSH,npMaGLSPH,tipJiangDDSH} mainly utilize the pairwise supervision or the triplet supervised information for hash learning, while ignoring the semantic class information which can help the improvement of the semantic discriminative ability of hash codes. Recently, some deep supervised hashing methods \cite{vcipMaMDLH,ijcaiYaoDSRH,nipsLiDSDH} improve the hashing retrieval performance by introducing the essential semantic structure of the data in form of class labels. \cite{ijcaiYaoDSRH} constructs a two-stream network architecture, one for classification stream and the other for hashing stream. However, the semantic labels do not directly guide the hash learning. \cite{nipsLiDSDH} assumes that the learned binary codes should be ideal for linear classification which restricts its scalability for some complex scene. \cite{vcipMaMDLH} weakens this assumpution to support nonlinear classification. However, \cite{vcipMaMDLH} utilizes the geometrical center of semantic relevant sub-centers as supervision information for multi-label hashing, which destroys the intrinsic manifold structure of the sub-center space.

In this paper, we propose a hierarchy neighborhood discriminative hashing method (HNDH). Specifically, we construct a bipartite graph to build coarse semantic neighbourhood relationship between the sub-class feature centers and the embeddings features. Moreover, we utilize the pairwise supervised information to construct the fined semantic neighbourhood relationship between embeddings features. Finally, we propose a hierarchy neighborhood discriminative hashing loss to unify the single-label and multi-label image retrieval problem with a one-stream deep neural network architecture. Compared with the multi-label classification loss in \cite{vcipMaMDLH} using the geometrical center of relevant sub-centers as supervision information, the proposed method employs the intrinsic manifold structure of these sub-centers for learning the discriminative hash codes. Some preliminary results on multi-task learning based semantic hashing framework were presented in \cite{vcipMaMDLH}, while the extension on the hierarchy neighborhood discriminative hashing loss and the unification of single-label and multi-label hash learning with one-stream network are novel. 

\begin{figure}[t]
	\centering
	\includegraphics[width=0.48\textwidth]{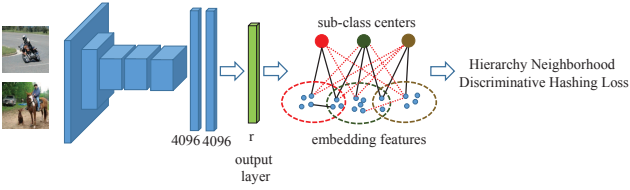} 
	\caption{An illustration of the proposed hierarchy neighborhood discriminative hashing method method (HNDH). The big filled circles with different colors denote the sub-class centers. The black solid lines connects two semantic relevant neighborhood vertexes. The red dashed lines connects two semantic irrelevant vertexes} 
	\label{fig:1} 
\end{figure}

\section{Hierarchy Neighborhood Discriminative Hashing}
\subsection{Problem Definition}
Assume we have a training set $X=\{x_{i}\}_{i=1}^{N}$, where $N$ denotes the number of training samples. The label information is denoted as $Y=\{y_{i}\}_{i=1}^{N}\in\{0,1\}^{L\times N}$, where $L$ denotes the number of categories. In addition, we can define a pairwise supervision matrix $S$ as $S_{ij}=1$ if $x_{i}$ and $x_{j}$ are semantically similar, and $S_{ij}=0$ otherwise. Under the supervised information $Y$ and $S$, supervised hashing aims to learn a hash function $h(\cdot)$ to transform the training data $X$ into a collection of $r$-bit compact binary codes $B=\{b_{i}\}_{i=1}^{N}\in\{-1,1\}^{r\times N}$. The Hamming distance between $b_{i}$ and $b_{j}$ is calculated by using $\text{dist}_{\mathcal{H}}=\frac{1}{2}(r-b^{T}_{i}b_{j})$. Therefore, we can utilize the inner product $\frac{1}{2}\langle b_{i}, b_{j}\rangle$ to measure the similarity of hash codes.

\subsection{Network Architecture}
As illustrated in Fig. \ref{fig:1}, the proposed one-stream deep architecture mainly contains two components: $1)$ the feature extraction subnetwork consists of the conv1 to fc7 layers of a pre-trained VGG-19 network \cite{nipsKrizhevskyVGG}; $2)$ the output layer for generating discriminative embedding features and hash codes for retrieval and classification. 

\subsection{Proposed Method}
\subsubsection{Coarse neighborhood discriminative hashing loss}
The output layer includes a $r$-dimensional fully-connected layer (4096-r) and a tangent layer to approximate the sign function. We utilize $u_{i}$ denotes the real-valued output of the network, i.e., the embedding features. For each sub-class, we compute its feature center as follows:
$c_{k}=\frac{1}{N_{k}}\sum_{i=1}^{N}Y_{ki}u_{i}$
where $c_{k}$ represents the centroid of the $k$-th sub-class, $N_{k}$ is the number of training samples that belong to the $k$-th sub-class. Therefore, we can construct a complete bipartite graph $G=(U,C,w)$ to build the relationship between the sub-class feature centers $C$ and the embeddings features $U$. The edge weight between the vertex $u_{i}$ and the vertex $c_{k}$ is defined as $w_{ik}=\frac{1}{2}c_{k}^{T}u_{i}$. Inspired by \cite{nipsGoldbergerNCA}, we can use a softmax normalization over the edge weights that connect the vertex $u_{i}$ to define the neighbour probability $p_{ik}$:
\begin{equation}
\begin{aligned}
p_{ik}=\frac{\text{exp}(\frac{1}{2}c_{k}^{T}u_{i})}{\sum_{k=1}^{L}\text{exp}(\frac{1}{2}c_{k}^{T}u_{i})}
\end{aligned}
\end{equation}
where $p_{ik}$ is the probability of $u_{i}$ selecting $c_{k}$ as its neighbor. The neighborhood relationship is coarse, since we only consider the relations between the embedding features and the sub-class centers. If the $k$-th sub-class is contained in the assigned labels of the embedding feature $u_{i}$, then the sub-class center $c_{k}$ is the relevant semantic neighborhood of $u_{i}$. Therefore, $c_{k}$ can participate in the class labels voting for $u_{i}$. Under this definition, the probability $p_{i}$ that image $i$ will be correctly classified can be computed as
\begin{equation}
\begin{aligned}
p_{i}=\sum_{k=1}^{L}Y_{ki}\frac{\text{exp}(\frac{1}{2}c_{k}^{T}u_{i})}{\sum_{k=1}^{L}\text{exp}(\frac{1}{2}c_{k}^{T}u_{i})}.
\end{aligned}
\end{equation}
The negative logarithm likelihood function can be defined as:
\begin{equation}
\begin{aligned}
J_{1}=-\sum_{i=1}^{m}\text{log}(\sum_{k=1}^{L}Y_{ki}\frac{\text{exp}(\frac{1}{2}c_{k}^{T}u_{i})}{\sum_{k=1}^{L}\text{exp}(\frac{1}{2}c_{k}^{T}u_{i})})
\end{aligned}
\end{equation}
where $m$ refers to the batch size. This function can minimize the intra-class variation and maximize the inter-class variation simultaneously to generate powerful embedding representations. It is worth noting that if $Y_{ki}$ is an one-hot vector, this function will become the single-label classification function in \cite{vcipMaMDLH,corrLiuCOCO}. However, this function is not restricted to single-label classification problem. It can extend to the multi-label classification problem naturally. Compared with the multi-label hashing loss in \cite{vcipMaMDLH} using the geometrical center of relevant semantic neighborhood sub-centers as supervision information, the proposed method employs the intrinsic manifold structure of the sub-center space for learning discriminative embedding features.

\subsubsection{Fined neighborhood discriminative hashing loss}
In the classification task above, we only consider the discrimination and polymerization of embedding features. The semantic neighborhood relationship between the embedding features is ignored which may make the distance between dissimilar embedding features smaller than the distance between similar embedding features. To overcome this limitation, we introduce the following pairwise constraint \cite{ijcaiLiDPSH} which is commonly used to preserve the semantic similarity in retrieval task.
\begin{equation}
\begin{aligned}
\label{eq:3}
J_{2} = -\sum_{i=1}^{m}\sum_{j=1}^{N}(S_{ij}\Theta_{ij}-\text{log}(1+e^{\Theta_{ij}}))
\end{aligned}
\end{equation}
where $\Theta_{ij}=\frac{1}{2}u_{i}^{T}u_{j}$. The semantic similarity matrix $S$ displays a fined neighborhood relationship between the embedding features, i.e., if $S_{ij}=1$ denotes that image $i$ and image $j$ are neighborhood in the semantic space and $S_{ij}=0$ otherwise. The neighborhood relationship is fined, since we consider the relations between all the embedding features.

\subsection{Objective Function and Learning Algorithm}
We formulate the proposed Hierarchy Neighborhood Discriminative Hashing method as the following multi-task learning framework:
\begin{equation}
\begin{aligned}
\label{eq:7}
J = NJ_{1} + \lambda J_{2} 
\end{aligned}
\end{equation}
where $N$ balance the impact of the different number of factors between the first term and the second term. We use an alternating optimization over the class sub-centers $C$ and the CNN parameters $\mathcal{N}$ as follows:
\begin{itemize}
	\item Fix $\mathcal{N}$ and optimize $C$. We can update the feature center of the $k$-th sub-class directly as follows:
	\begin{equation}
	\begin{aligned}
	\label{eq:8}
	c_{k}=\frac{1}{N_{k}}\sum_{i=1}^{N}Y_{ki}u_{i}
	\end{aligned}
	\end{equation}
	\item Fix $C$ and optimize $\mathcal{N}$. 
	\begin{equation}
	\begin{aligned}
	\label{eq:9}
	\frac{\partial J}{\partial u_{i}} = &\frac{N}{2}(P_{*i}-Q_{*i})^{T}C+\frac{\lambda}{2}\sum_{j:S_{ij}\in S}(a_{ij}-S_{ij})u_{j} \\
	&+\frac{\lambda}{2}\sum_{j:S_{ji}\in S}(a_{ji}-S_{ji})u_{j}
	\end{aligned}
	\end{equation}
	where $P_{mi}=\frac{\text{exp}(\frac{1}{2}c_{m}^{T}u_{i})}{\sum_{m=1}^{L}\text{exp}(\frac{1}{2}c_{m}^{T}u_{i})}$, $Q_{mi}=\frac{Y_{km}\text{exp}(\frac{1}{2}c_{m}^{T}u_{i})}{\sum_{m=1}^{L}Y_{km}\text{exp}(\frac{1}{2}c_{m}^{T}u_{i})}$, $C=[c_{1},c_{2},\cdots,c_{L}]$ and $a_{ij}=\sigma(\Omega_{ij})$. Then we can compute $\frac{\partial J}{\partial \Theta}$ with $\frac{\partial J}{\partial u_{i}}$ by using the chain rule. In each iteration, we update the parameter $\Theta$ based on the backpropagation (BP) algorithm.
\end{itemize}

\setcounter{table}{0}
\begin{table}[!htbp]
	\centering
	\small
	\caption{Statistics of the datasets used in the experiments.}
	\label{table:1}
	\setlength{\tabcolsep}{1.8mm}{
		\begin{tabular}{|c|c|c|c|}
			\hline
			Dataset & Total & Query / Retrieval / Training & Labels \\ \hline
			CIFAR-10 & 60,000 & 1,000 / 5,9000 / 5,000 & 10 \\ \hline
			NUS-WIDE & 195,834 & 2,100 / 193,734 / 10,500 & 21\\
			\hline
		\end{tabular}
	}
\end{table}

\section{Experimental Results}
\subsection{Datasets and Experimental Settings}
We evaluate the performance of several deep hashing methods on two public datasets: CIFAR-10 and NUS-WIDE. We split each dataset into a query set and a retrieval set. The training set is randomly selected from the retrieval set. For the CIFAR-10 dataset, 100 images per class are randomly selected as the query set and the remaining images are used as the retrieval set following \cite{ijcaiLiDPSH,accvWangDTSH,nipsLiDSDH}. Moreover, 500 images per class are randomly sampled from the retrieval set as the training set. For the NUS-WIDE dataset, we only use the images that belong to the 21 most frequent labels. Then it contains at least 5,000 images for each class. We randomly sample 2100 images (100 images per class) as the query set and the remaining images form the retrieval set. Moreover, 500 images per class from the retrieval set are used as training set. The statistics of the two dataset splits are summarized in Table \ref{table:1}. For CIFAR-10, we use Mean Average Precision (MAP) as the evaluation metric following \cite{ijcaiLiDPSH,accvWangDTSH,nipsLiDSDH}. The MAP@5K for NUS-WIDE is evaluated on top 5,000 retrieved images as similar in \cite{ijcaiLiDPSH,accvWangDTSH,nipsLiDSDH}.


We use two NVIDIA TITAN XP GPUs and MatConvnet as the platform to implement the proposed model. The pre-trained VGG-19 model is utilized to initialize the base network in HMDH and the other parameters of the network are randomly initialized. The iteration number of the proposed HMDH is set to be 100 and the batch size is fixed to 128 for all datasets. The learning rate of the base network is gradually reduced from $10^{-2}$ to $10^{-3}$ for both the CIFAR-10 and NUS-WIDE datasets. The learning rate for the newly added layers is set to be 10 times more than the layers of the base network. For both datasets, we set $\lambda=1$ via cross validation on training sets.

\begin{table}[!thbp]
	\centering
	\normalsize
	\caption{The comparisons of MAP on \textbf{CIFAR-10} dataset.}
	\label{table:2}
	\begin{tabular}{|c|c|c|c|c|}
		\hline
		\multirow{2}{*}{Method}&\multicolumn{4}{c|}{\textbf{CIFAR-10}}   \cr\cline{2-5}
		\multirow{2}{*}{}&12 bits&24 bits&32 bits&48 bits \cr\cline{2-5}
		\hline
		\hline
		HNDH &\textbf{0.805}&0.825&0.829&\textbf{0.838}\cr\cline{1-5}
		MLDH \cite{vcipMaMDLH} &\textbf{0.805}&0.825&0.829&\textbf{0.838}\cr\cline{1-5}
		DDSH \cite{tipJiangDDSH} &0.769 &\textbf{0.829} &\textbf{0.835} &0.819 \cr\cline{1-5}
		DSDH \cite{nipsLiDSDH} &0.740&0.786&0.801&0.820\cr\cline{1-5}
		DTSH \cite{accvWangDTSH} &0.710&0.750&0.765&0.774\cr\cline{1-5}
		DPSH \cite{ijcaiLiDPSH}&0.713&0.727&0.744&0.757\cr\hline
	\end{tabular}
\end{table}

\begin{table}[!thbp]
	\centering
	\normalsize
	\caption{The comparisons of MAP@5K on \textbf{NUS-WIDE} dataset.}
	\label{table:3}
	\begin{tabular}{|c|c|c|c|c|}
		\hline
		\multirow{2}{*}{Method}&\multicolumn{4}{c|}{\textbf{NUS-WIDE}} \cr\cline{2-5}
		\multirow{2}{*}{}&12 bits&24 bits&32 bits&48 bits \cr\cline{2-5}
		\hline
		\hline
		HNDH &\textbf{0.806}&\textbf{0.832}&\textbf{0.841}&\textbf{0.848}\cr\cline{1-5}
		MLDH \cite{vcipMaMDLH} &0.800&0.828&0.832&0.835\cr\cline{1-5}
		DDSH \cite{tipJiangDDSH} &0.791 &0.815 &0.821 &0.827 \cr\cline{1-5}
		DSDH \cite{nipsLiDSDH} &0.776&0.808&0.820&0.829\cr\cline{1-5}
		DTSH \cite{accvWangDTSH} &0.773&0.808&0.812&0.824\cr\cline{1-5}
		DPSH \cite{ijcaiLiDPSH}&0.752&0.790&0.794&0.812\cr\hline
	\end{tabular}
\end{table}

\begin{table*}[!thbp]
	\centering
	\normalsize
	\caption{The impact of different components of our HNDH on MAPs for \textbf{CIFAR-10} and \textbf{NUS-WIDE} datasets.}
	\label{table:4}
	\begin{tabular}{|cc|c|c|c|c|c|c|c|c|}
		\hline
		\multicolumn{2}{|c|}{Method}&\multicolumn{4}{c|}{\textbf{CIFAR-10}}  &\multicolumn{4}{c|}{\textbf{NUS-WIDE}} \cr\cline{1-10}
		$J_{1}$&$J_{2}$&12 bits&24 bits&32 bits&48 bits &12 bits&24 bits&32 bits&48 bits \cr\cline{1-10}
		\hline
		\hline
		$\surd$&$\surd$&\textbf{0.8045}&\textbf{0.8250}&\textbf{0.8293}&\textbf{0.8377}&\textbf{0.8062}&\textbf{0.8317}&\textbf{0.8410}&\textbf{0.8484}\cr\hline
		&$\surd$&0.7524&0.7865&0.7883&0.7888&0.7860&0.8204&0.8270&0.8339\cr\cline{3-10}
		$\surd$&       &0.7640&0.8085&0.8125&0.7977&0.7594&0.8007&0.8102&0.8186\cr\hline
	\end{tabular}
\end{table*}

\begin{figure*}[!thbp]
	\centering
	\subfloat[HNDH-F]{
		\includegraphics[width=.22\textwidth]{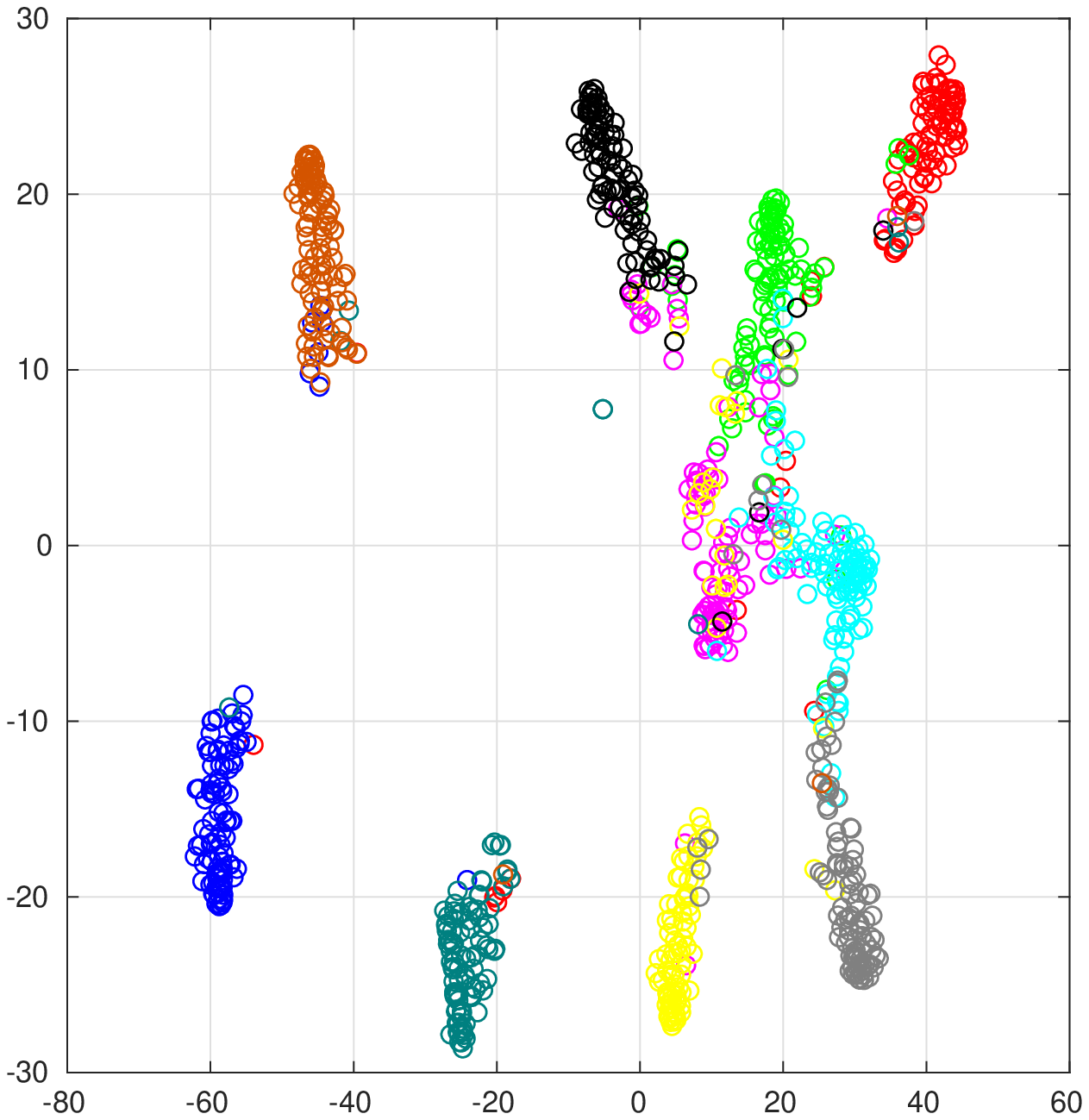}}
	\subfloat[HNDH-C]{
		\includegraphics[width=.22\textwidth]{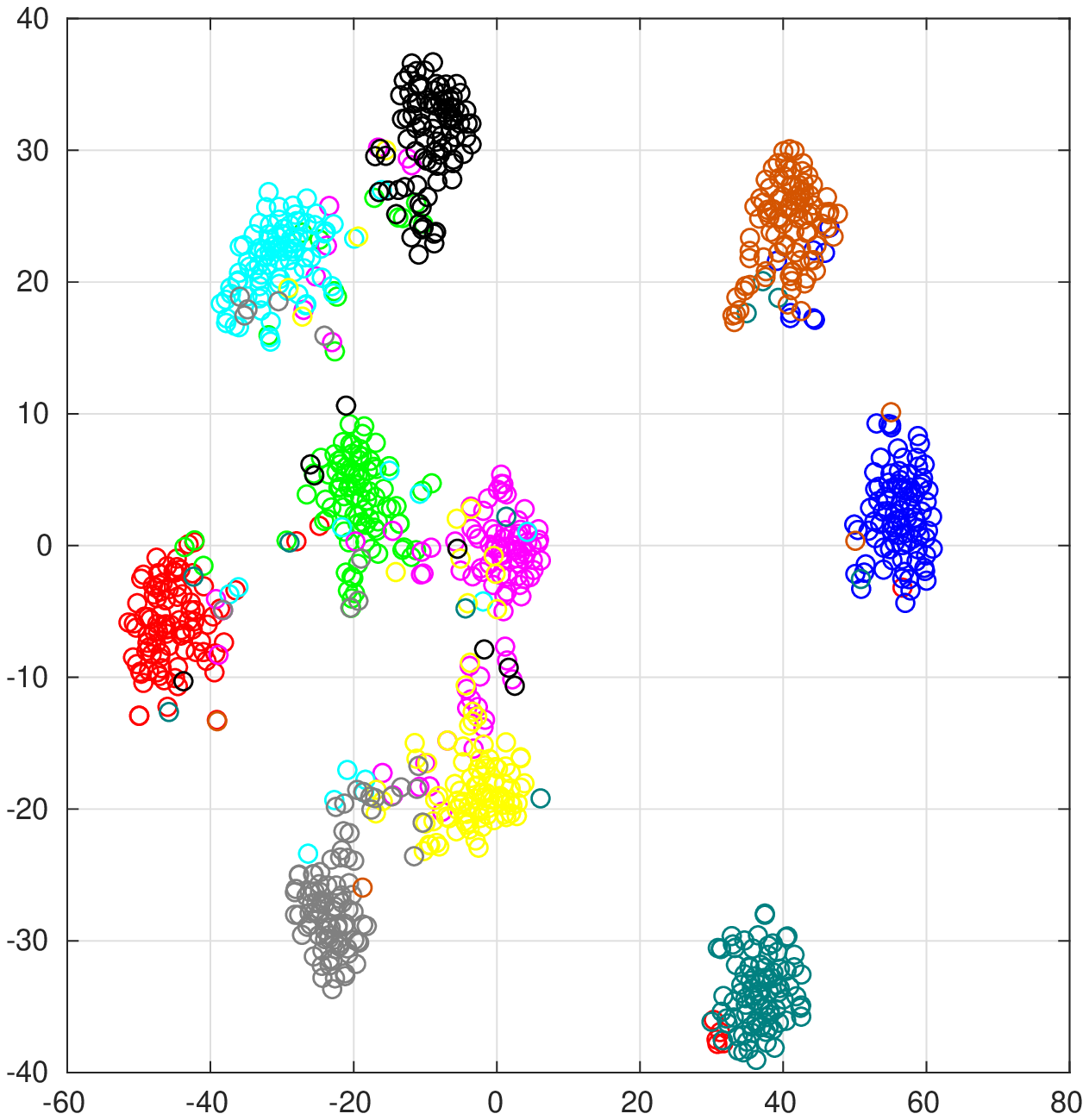}}
	\subfloat[HNDH]{
		\includegraphics[width=.22\textwidth]{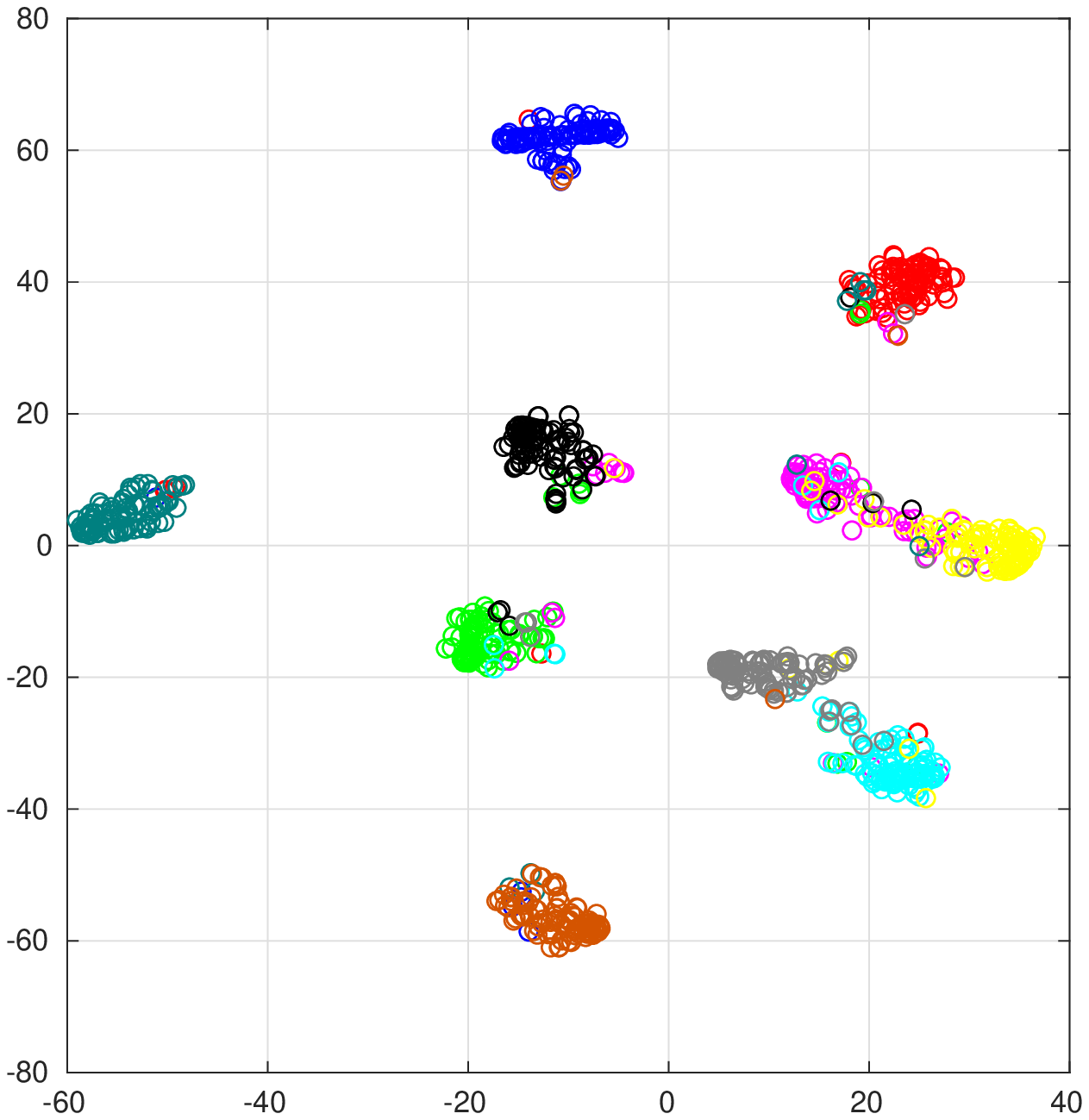}}
	\subfloat[Training curves]{
		\includegraphics[width=.22\textwidth]{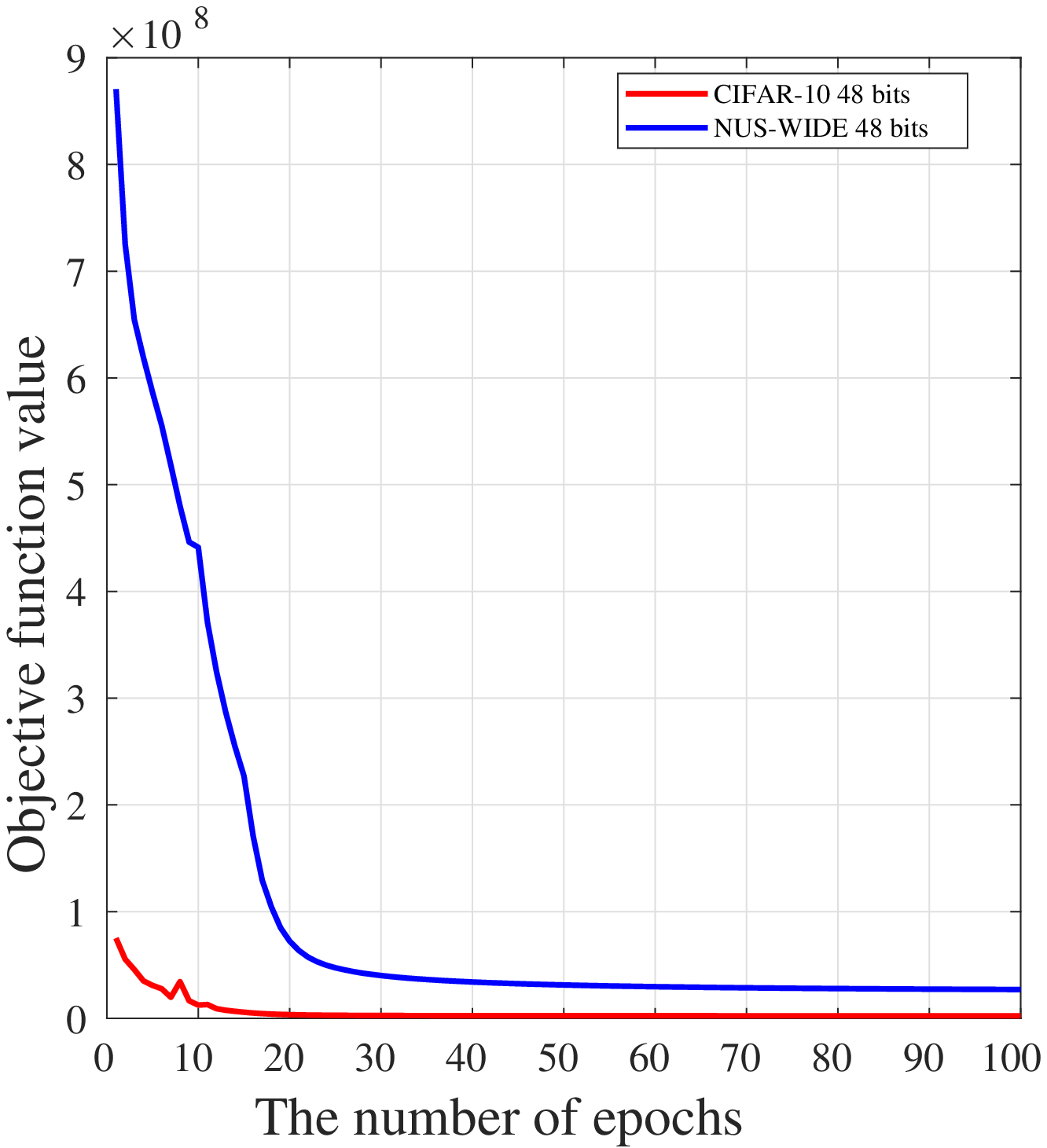}}
	\caption{$(a)\sim(c)$ show the t-SNE visualization of the deep representations of HNDH-F, HNDH-C, and HNDH with 48 bits on the query set of CIFAR-10 dataset. (d) shows training convergence curves of the proposed model at 48 bits over CIFAR-10 and NUS-WIDE datasets.}
	\label{fig:2}
\end{figure*}

\subsection{Results and Discussions}
The MAP results of CIFAR-10 and NUS-WIDE are presented in Table \ref{table:2} and Table \ref{table:3} respectively. The results of deep supervised baselines including \cite{tipJiangDDSH,ijcaiLiDPSH,accvWangDTSH} and \cite{tipJiangDDSH} on CIFAR-10 and NUS-WIDE are cited from \cite{nipsLiDSDH} and \cite{tipJiangDDSH} respectively. It can be seen that the proposed method outperforms the other baselines for most cases. The average MAP of the proposed method is 0.824, which is 1.1 percents higher than the average of DDSH's 0.813 on the CIFAR-10 dataset. On the NUS-WIDE dataset, the proposed method performs consistently better than the other baselines across all bits. The average MAP@5K of the proposed method is 0.831, which is 0.7 percents higher than the average of MDLH's 0.824 on the NUS-WIDE dataset. The reason can be that the relations between the learned embedding feature from MDLH tends to locate at geometrical center of its semantic relevant sub-class centers which destroys the manifold structure in the sub-center space. When the hash code is short (e.g., 16 bits), the proposed method and MDLH perform much better than the state-of-the-art. The reason is that the semantic label information is employed to learn the discrimination and polymerization of hash codes. Although \cite{nipsLiDSDH} also utilizes the label information, they ignore the polymerization of hash codes.

Compared to the single-task learning based hashing methods including DTSH \cite{accvWangDTSH}, DPSH \cite{ijcaiLiDPSH}, HashNet \cite{iccvCaoHashNet}, DHN \cite{aaaiCaoDHN}, DNNH \cite{cvprLaiDNNH} and CNNH \cite{aaaiXiaCNNH}, the multi-task learning based hashing methods including the proposed method, MLDH \cite{vcipMaMDLH} and DSDH \cite{nipsLiDSDH} jointly consider the the retrieval task and the classification task for learning the discrete discriminative hash codes. From Table \ref{table:2} and Table \ref{table:3}, it can be found that the multi-task learning based hashing methods generally perform better than the single-task learning based hashing methods. In addition, different from DSDH, the polymerization of hash codes is also considered in the proposed method.


\subsection{Ablation Experiments}
We report the effect of different components of our HNDH on two benchmark datasets with different numbers of bits in Table \ref{table:4}. From the MAP results, it verifies the effectiveness of combining the coarse neighborhood discriminative hashing loss and the fined neighborhood discriminative hashing loss. In addition, the proposed the individual coarse neighborhood discriminative hashing loss $J_{1}$ performs better than the individual multi-label hashing loss in \cite{vcipMaMDLH}. To facilitate the outstanding, we focus on two HNDH variants: (a) HNDH-C is the first variant which removes fined neighborhood discriminative hashing loss; (b) HNDH-F is the second variant which removes coarse neighborhood discriminative hashing loss. Fig. \ref{fig:2} $(a)\sim(c)$ show t-SNE visualization \cite{JMLRvantSNE} of the deep representations of HNDH-F, HNDH-C, and HNDH with 48 bits on the query set of CIFAR-10 dataset. As shown in Fig. \ref{fig:2} $(a)\sim(c)$, the image embeddings generated by HNDH show most compact and discriminative structures with clearest boundaries.

\subsection{Convergence Analysis}
The training convergence curves of the proposed model at 48 bits over CIFAR-10 and NUS-WIDE datasets are shown in Fig. \ref{fig:2} (d). It can be observed that the proposed model can converge within 100 iterations, which validates the effectiveness of the proposed approach.

\section{Conclusion}
In this paper, we propose a hierarchy neighborhood discriminative hashing method. Firstly, we construct a bipartite graph to build coarse semantic neighbourhood relationship between the sub-class feature centers and the embeddings features. Moreover, we utilize the pairwise supervised information to construct the fined semantic neighbourhood relationship between embeddings features. Finally, we propose a hierarchy neighborhood discriminative hashing loss to unify the single-label and multi-label image retrieval problem with a one-stream deep neural network architecture. In the future work, we plan to extend the proposed single-modal hashing method to the cross-modal hashing.

\bibliographystyle{abbrv}
\bibliography{references}
\end{document}